%% file: root.tex
\documentclass[letterpaper, 10 pt, conference]{ieeeconf}  

\IEEEoverridecommandlockouts                              

\input{macros}

\title{\LARGE \bf
Transfer-LMR: Heavy-Tail Driving Behavior Recognition in Diverse Traffic Scenarios
}

\author{Chirag Parikh$^{1}$, Ravi Shankar Mishra$^{1}$, Rohan Chandra$^{2}$, Ravi Kiran Sarvadevabhatla$^{1}$ 
\thanks{$^{1}$Centre for Visual Information Technology, IIIT Hyderabad, India {\tt\small chirag.parikh@research.iiit.ac.in}}%
\thanks{$^{2}$University of Texas at Austin, USA
        {\tt\small rchandra@utexas.edu}}%
}

\begin{document}

\maketitle
\thispagestyle{empty}
\pagestyle{empty}

\begin{abstract}
Recognizing driving behaviors is important for downstream tasks such as reasoning, planning, and navigation. Existing video recognition approaches work well for common behaviors (e.g. “drive straight”, “brake”, “turn left/right”). However, the performance is sub-par for underrepresented/rare behaviors typically found in tail of the behavior class distribution. To address this shortcoming, we propose Transfer-LMR, a modular training routine for improving the recognition performance across all driving behavior classes. We extensively evaluate our approach on METEOR and HDD datasets that contain rich yet heavy-tailed distribution of driving behaviors and span diverse traffic scenarios. The experimental results demonstrate the efficacy of our approach, especially for recognizing underrepresented/rare driving behaviors.


\end{abstract}


\section{Introduction}


Driving behavior of a vehicle is a combination of driving micro-actions (accelerate, brake, turn, etc.) accompanied by unique long-horizon maneuvers executed in specific road and traffic conditions. In real-world traffic scenarios, certain behaviors are rarely observed as compared to others. Due to this, the naturally collected driving datasets contain a heavily tailed distribution of unique driving situations \cite{koopman2018heavy,chandra2023meteor,ramanishka2018toward}. 
Tackling the heavy-tail problem by effectively recognizing all unique driving behaviors in different traffic situations is crucial for achieving a high level of safety with autonomous vehicles~\cite{koopman2018heavy, chandra2019densepeds, chandra2020roadtrack}. This enables robust path planning and navigation in diverse traffic scenarios~\cite{chandra2022towards,ramanishka2018toward,xu2019temporal,shimosaka2014modeling, chandra2019traphic, chandra2019robusttp, chandra2020forecasting}. 


While multiple approaches exist for recognizing driving behaviors from dashcam videos \cite{li2020learning} \cite{noguchi2023ego}, they have primarily been designed to work well for balanced dataset distributions. Due to the natural heavy-tail distribution of driving behavior datasets, they exhibit subpar class-wise average recognition performance. Specifically, the recognition performance of the tail class behaviors is overshadowed by similar head class behaviors.
While multiple methods exist for modeling the imbalanced dataset distribution, they are primarily based on large-scale video-recognition datasets that contain a huge variety of human-action class categories (100+) \cite{zhang2021videolt,perrett2023use,li2023meid,moon2023minority}. Driving behavior datasets are comparatively low-resource and contain very few unique class types ($<12$) \cite{chandra2023meteor,ramanishka2018toward}. Due to the limited variety of unique classes, existing long-tailed methods are not directly applicable to driving datasets with heavy-tailed distribution.

\begin{figure}
    \centering
    \includegraphics[width=\columnwidth]{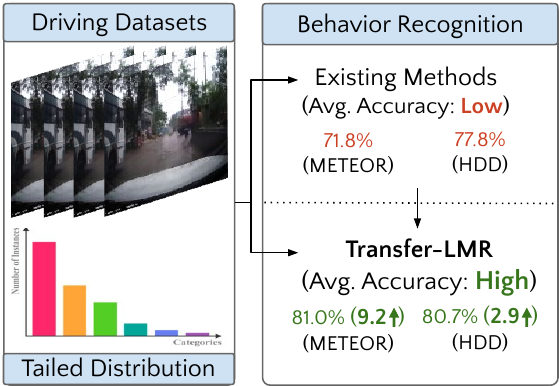}
    \caption{Heay-Tail Distributed ego-vehicle driving behavior recognition in diverse traffic environments. Our proposed approach, Transfer-LMR, significantly outperforms existing behavior recognition methods by 9.2\% and 2.9\% in average class accuracy on real-world driving datasets METEOR\cite{chandra2023meteor} and HDD\cite{ramanishka2018toward}.}
    \label{fig: cover}
    \vspace{-10pt}
\end{figure}
%

To improve the representation of the tail classes in low-resource driving datasets, we introduce an approach, Transfer-LMR, for heavy-tail driving behavior recognition in diverse traffic environments. Transfer-LMR is a multi-stage training routine with a video feature refinement technique for mitigating the class imbalance problem by utilizing the similarity of features between the head and the tail behavior classes. An overview of our approach is shown in Figure \ref{fig: cover}.



\subsection{Main Contributions} 
\begin{enumerate}

\item We contribute Transfer-LMR, the first approach for tackling heavy-tailed driving behavior recognition using low-resource dashcam-video datasets captured in diverse traffic environments.

\item We establish state-of-the-art performance on the METEOR~\cite{chandra2023meteor} and HDD~\cite{ramanishka2018toward} datasets using our approach, boosting average class accuracy by 9.2\% and 2.9\% respectively without sacrificing the overall performance (mAP) compared to existing baselines. 

\item Through Transfer-LMR, we provide a modular plug-and-play training recipe showcasing its applicability to diverse video backbones (e.g. C3D, I3D, MotionFormer) while also being fast operating at $2-3$ milliseconds per video ($64$ frames).



\end{enumerate}

We showcase the efficacy and generalizability of our approach by benchmarking on two publicly-available driving video datasets \cite{chandra2023meteor} \cite{ramanishka2018toward} with tactical and atypical driving behaviors. We establish the benchmarks using different video backbones \cite{tran2015learning} \cite{carreira2017quo} \cite{patrick2021keeping} and show significant improvements over the baselines in average class accuracy (Avg. C/A) while also improving the overall mean average precision (mAP). A video overview of our approach can be found in the Supplementary material.



\section{Related Work}

In this section, we provide an overview of the existing literature on driving behavior recognition and long-tailed video activity recognition. 

\subsection{Ego Vehicle Driving Behavior Recognition}
Multiple research works~\cite{chandra2021using, chandra2020cmetric, chandra2020graphrqi, mavrogiannis2022b} have explored the use of sensor data for recognizing normal \cite{oliver2000graphical} \cite{mitrovic2005reliable}, aggressive \cite{ma2018comparative}, and anomalous \cite{matousek2018robust} \cite{matousek2019detecting} ego vehicle driving behaviors. However, it is often infeasible to differentiate between certain behaviors using sensor data when the same driving action is performed in different situations. For example, ego vehicle deviating on valid driving lanes vs. when it deviates towards an invalid or wrong lane has the same deviating driving action but is performed in different road conditions. 

Robust recognition of such driving behaviors (atypical \cite{chandra2023meteor} and tactical \cite{ramanishka2018toward}) requires strong visual cues. Some of the proposed approaches use front-view dashcam videos along with sensor data \cite{xu2017end} \cite{ramanishka2018toward}. While recently proposed approaches solely rely on dashcam videos \cite{li2020learning} \cite{noguchi2023ego}, Li et al. \cite{li2020learning} proposed the use of Graph convolution networks (GCN) for modeling egocentric spatial-temporal
interaction with road agents and infrastructure from input videos. Noguchi et al. \cite{noguchi2023ego} incorporated a semi-supervised contrastive learning technique and utilized unlabelled video data for improving the overall performance of driving behavior recognition. However, due to the heavy-tailed distribution of behavior labels in driving datasets, existing approaches get biased towards accurately predicting the head classes while overshadowing the tail classes and degrading the class-wise average performance across the entire dataset. Our proposed approach primarily focuses on solving this problem without sacrificing the model's overall performance on the dataset.

\subsection{Long-Tailed Recognition}
Unlike driving datasets, large-scale human-action video and image datasets have a long-tailed distribution with many head and tail classes. Recently, there have been increasing efforts for long-tailed video recognition \cite{zhang2021videolt,perrett2023use,li2023meid,moon2023minority} following the image-based techniques \cite{kang2019decoupling,zhang2021distribution,alshammari2022long,tang2022invariant} for such large-scale datasets. Zhang et al. \cite{zhang2021videolt} propose FrameStack which adaptively samples different numbers of frames from each video based on running average precision for each class. Li et al. \cite{li2023meid} consider frame-level imbalance with frame-level feature learning. Moon et al. \cite{moon2023minority} simultaneously handle data imbalance at the video level through a combination of frame-wise learnable feature aggregators and vicinity expansion based on class frequency alleviating majority-biased training. \cite{perrett2023use} jointly reconstructs the features of all samples in the batch weighted by the class count. However, these methods are not designed for low-resource driving video datasets that have a limited variety of head and tail classes. Our proposed Transfer-LMR approach is specifically adapted to alleviate the problems due to the heavy-tailed distribution and the limited variety of classes in driving datasets.

\section{Background}
\subsection{Problem Formulation}
\label{sub: Bg_sampling}


Approaches for video-based driving behavior recognition combine a backbone for extracting features $Z$ and a classifier network $g$ for class prediction. During training, the model receives the data samples in mini-batches, denoted as $X$, comprising $B$ videos, where $X = \{x_i : i = 1 \ldots B\}$, and their corresponding labels $Y = \{y_i : i = 1\ldots B\}$. Every video sample $x_i$ is augmented using standard methods (eg. random-resized cropping, temporal jittering, etc.) before feeding it as input to the backbone network. Standard cross-entropy loss function is used for training the entire model and is given by:
\begin{equation} 
\label{eq:CE}
L_{CE}(Y, \hat{Y}) = -\frac{1}{B} \sum_{i=1}^{B} \sum_{j=1}^{C} y_{ij} \cdot \log(\hat{y}_{ij})
\end{equation}
where $\hat{Y}$ corresponds to the predicted labels and $C$ is the number of unique classes in the dataset.


\begin{figure*}[t]
    \centering
    \includegraphics[width=\textwidth]{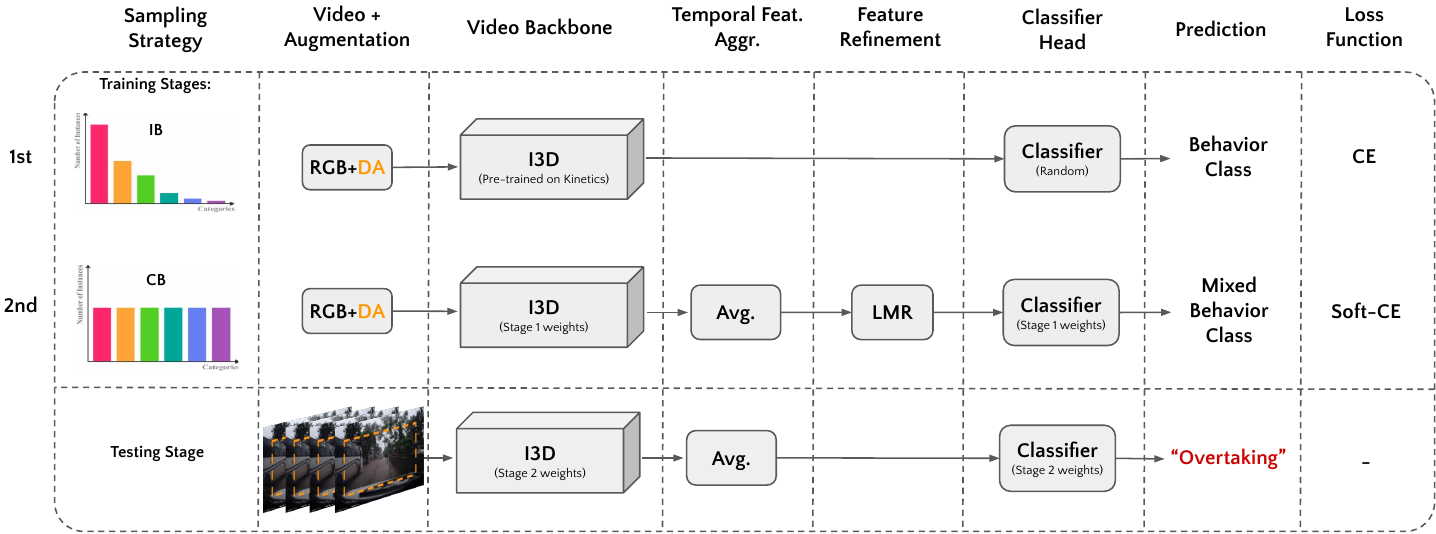}
    \caption{\textbf{TransferLMR:} A Two-Stage Training Method for Heavy-Tail Driving Behavior Recognition. Instance-balanced (IB) and class-balanced (CB) are sampling strategies. DA is our customized data augmentation method. Averaging (Avg.) is used for Temporal Feature Aggregation. Long-tailed Mixed Reconstruction (LMR) is used for Feature Refinement. Driving Behavior Class prediction in the Testing Stage does not require LMR.}
    \label{fig: approach_diagram}
\end{figure*}

\textbf{Sampling Strategies.} To form the mini-batch, data points $(x_i, y_i)$ are randomly sampled from the training set with different strategies, namely instance-balanced (IB), and class-balanced (CB). With the IB method, all data points get equal opportunity of being picked up during training. Formally its data sampling probability, $p_j^{IB}$ for class $j$, is directly proportional to the total number of instances $n_j$ from that class and is given by:
\begin{equation}
    p_j^{IB} = \frac{n_j}{\sum_{i=1}^C n_i} 
\label{eq: sampling}
\end{equation}

This is the most common sampling method and is used by existing behavior recognition approaches~\cite{li2020learning,noguchi2023ego}.

The CB method samples data points from every behavior class in equal proportions to form the mini-batch. It's data sampling probability $p_j^{CB}$ for every class $j$ is $1/C$. The CB method is more suitable for imbalanced or heavy-tail distributed datasets to enable learning a balanced representation towards all classes during training. 

\subsection{Long-Tailed Mixed Reconstruction (LMR)}
\label{sub: LMR}

During training the data points from tail class are oversampled when using the class-balanced sampling strategy. This causes the model to overfit the limited diversity of the tail class samples in the training dataset. To alleviate this issue, Perrett et al. \cite{perrett2023use} proposed the LMR approach that creates hybrid features of training data points using feature reconstruction and pairwise mixing to increase the tail class diversity. 




\textbf{Sample reconstruction.} Features $Z_i$ of each video sample $x_i$ is reconstructed using a combination of similar features in the mini-batch. The similarity between features is calculated using the cosine metric given by $S_{ij} = s(Z_i, Z_j)$, where $j \neq i$. The feature combinations are weighted by $W_{ij}$ that is calculated using the softmax operation applied over $S_{ij}$ and is given by:

\begin{equation} 
\label{eq:softmax}
W_{ij} = \frac{\exp(S_{ij})}{\sum_{k=1}^{B} \exp(S_{ik})} \text{, where } j, k \neq i
\end{equation}

The reconstructed features $R_i$ are then obtained by:
\begin{equation}
    R_i = \sum_{j=1}^{B} W_{ij} Z_j \text{, where } j \neq i
\end{equation}

Finally, the original features $Z_i$ are fused with the reconstructed features $R_i$ using a class contribution function $\mathbf{c}(y_i)$, as defined in~\cite{perrett2023use}, to obtain the features $M_i$ given by:

\begin{equation} \label{eq:contribution_rep_rec}
M_i = \mathbf{c}(y_i) R_i + (1-\mathbf{c}(y_i))Z_i
\end{equation}








These features $M_i$ and their labels $y_i$ are then updated through stochastic pairwise mixing as defined in \cite{perrett2023use} to obtain mixed features and mixed labels that helps in learning robust decision boundaries. The entire LMR approach focuses on improving the tail behavior class representation without losing the quality of the head class features. 




\section{Transfer-LMR}
\label{sec: methodology}



The training practices followed by the LMR approach are suitable for large-scale and long-tail distributed human-action datasets. They are not directly applicable to low-resource and heavy-tail distributed driving video datasets. This is because its effectiveness relies on the quality of feature representations learned in the pre-training stage. These learned representations are of substandard quality for driving datasets due to their inferior scale and limited label variety as compared to human-action video datasets. We propose Transfer-LMR, an approach for systematically obtaining high-quality feature representations followed by their successive refinement using LMR to increase the tail behavior class diversity for low-resource driving datasets. This is achieved through a two-stage training pipeline that includes different data sampling strategies, a customized data augmentation method to preserve the peripheral image context, temporal aggregation of video features, and the unique application of LMR for its refinement. Our approach improves the representation of the tail behavior classes and boosts the class-wise average recognition performance on driving datasets.

In the following subsections, we first describe the customized data augmentation method incorporated during the training stages. Then we elaborate on the two-stage training pipeline, LMR-based feature refinement, and the temporal feature aggregation methods employed in our approach.

\subsection{Data Augmentation}
\label{sub: D_S_A}
\begin{figure}
    \centering
    \includegraphics[width=\columnwidth]{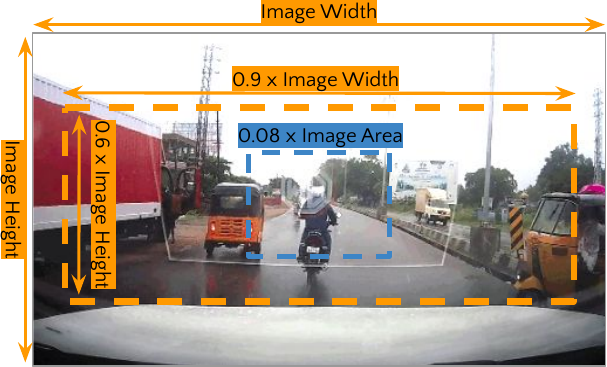}
    \caption{Modified random resized cropping augmentation method tailored for dashcam-based driving videos. The dotted orange line represents the position and relative dimensions of the minimum cropping region obtained from this method, while the dotted blue line corresponds to that obtained from the original unaltered augmentation method.}
    \label{fig: DataAugmentation}
\end{figure}

Standard random resized cropping augmentation, used for human-action video datasets, arbitrarily chooses a cropping region from anywhere inside the image boundary. But, front-view dashcam videos often contain important road and traffic objects around the image periphery that influence the ego vehicle's driving behavior. These objects may get cropped out after applying the standard augmentation method. Thus to preserve such peripheral context, we modify the standard augmentation method to restrict cropping around the left and right image boundaries. We implement this by empirically defining the minimum cropping width and height ratios $r_w$ and $r_h$ as 0.9 and 0.6 respectively of the original image. Based on this the aspect ratio range for the crop bounding box is given by  $A_{min}$ and $A_{max}$ as: 

\begin{equation}
    A_{min} = \frac{r_w \cdot w}{h}
\end{equation}
\begin{equation}
    A_{max} = \frac{w}{r_h \cdot h}
\end{equation}

where $w$ and $h$ are image width and height respectively. The minimum cropping area ratio using our modified method is given by $AR_{min} = r_w r_h$, i.e., 0.54 while that for the standard augmentation method is 0.08. A demonstration of the minimum cropping regions chosen using our modified method and the standard method is shown in Figure \ref{fig: DataAugmentation} with orange and blue dotted lines respectively. We employ this modified augmentation method on the input dashcam videos during both the training stages of our approach described in the following subsection.

\subsection{Two-Stage Training Routine}
\label{sub: twostages}

The two-stage training methodology of Transfer-LMR is demonstrated in Figure \ref{fig: approach_diagram}. 

\textbf{First Stage.} The first training stage comprises a video backbone pre-trained on a large-scale video recognition dataset and a classifier network initialized from scratch for driving behavior class prediction. During this stage, the entire model is trained end-to-end with instance-balanced data sampling (Section~\ref{sub: Bg_sampling}) and the sampled driving videos are augmented with our customized boundary-constrained cropping method (Section~\ref{sub: D_S_A}). A standard cross-entropy loss function is used to optimize for the driving behavior recognition task. This stage ensures the transfer of high-quality video features to the driving domain yielding superior overall behavior recognition performance.


\textbf{Second Stage.} Due to the heavy tail distribution of ego vehicle's driving behavior categories, the model obtained from the first training stage underfits the tail behavior classes thereby giving subpar class-wise average recognition performance\cite{li2020learning} \cite{noguchi2023ego}. The second training stage is primarily meant to alleviate this issue by refining the high-quality video features obtained from the first training stage. In this stage, we first sample the training data points using the CB sampling strategy for rebalancing the data distribution (Section~\ref{sub: Bg_sampling}). Secondly, we employ the modified data augmentation method, which was also used in the first training stage, to preserve the peripheral context from the input video's RGB frames. Next, we feed the augmented RGB frames to the I3D backbone \cite{carreira2017quo} and extract the video features for its successive refinement using the LMR approach (Section \ref{sub: LMR}).

The LMR approach is suited for refining video features that are in condensed form having no temporal dimension. However, the video features, from the I3D backbone, have an additional temporal dimension and thus require aggregation. The temporal feature aggregation is accomplished by simple averaging across the temporal dimension and is given by the following equation:
\begin{equation} 
\label{eq:temporal_avg}
Z_{i} = \frac{\sum_{t=1}^{T} Z_{it}}{T} 
\end{equation}
where $Z_{it}$ is the feature at temporal index $t$, and $T$ is the total number of temporal features. Alternative to this approach, we also adopted learnable feature aggregation methods including Bi-directional GRU and Transformer but they were found to yield subpar performance in comparison to simple averaging. This indicated that learnable aggregation methods are not well-suited for low-resource driving video datasets. 

The LMR module consumes the aggregated features and their corresponding ground-truth behavior labels $Y$ and produces refined features and their corresponding ground-truth mixed behavior labels $Y^*$ using stochastic pairwise mixing as defined in \cite{perrett2023use}. The refined features are then fed to the classifier network to yield the predicted labels $\hat{Y}$ as the model's output. The mixed behavior labels $Y^*$ and the predicted labels $\hat{Y}$ are optimized using the soft cross-entropy loss function $L_{SCE}(\cdot, \cdot)$ given by:
\begin{equation} 
\label{eq:SCE}
L_{SCE}(Y^*, \hat{Y}) = -\frac{1}{B} \sum_{i=1}^{B} \sum_{j=1}^{C} {y}_{ij}^* \cdot \log(\hat{y}_{ij})
\end{equation}
It is to be noted that we reuse the classifier network weights that were obtained from the first training stage as opposed to the original LMR approach and also aggregate the temporal features instead of its independent application across the time dimension. 



The entire network is fine-tuned with a reduced learning rate and a tuned set of hyperparameters, as defined in \cite{perrett2023use}, for the LMR feature refinement module. During the testing or inference stage, the LMR-based feature refinement technique is disabled and the temporally aggregated and refined video features extracted from an input video are directly fed to the classifier network for driving behavior prediction. This two-stage training procedure thus increases the diversity of the under-represented behavior classes and helps in obtaining better class-wise average recognition performance as compared to the baselines while also maintaining a better overall performance.

\begin{table*}[!ht]
    \begin{center}
    \begin{threeparttable}
    
    \caption{Driving Behavior Recognition on \textbf{METEOR dataset \cite{chandra2023meteor}}. Column TFA corresponds to Temporal Feature Aggregation methods. Abbreviations for driving behavior category: Overtaking (OT), Deviating on Unmarked Lanes (DT), Rule Breaking by Wrong Lane driving (WL), Cutting (CT), Yielding (YD). The frequency of each behavior class in the training set is specified in round brackets alongside its abbreviation. Color shades from yellow to orange indicate the heavy-tailed distribution of driving behavior categories in the dataset.}
     
    \label{tab:METEORresults}
    
    \scriptsize
    \renewcommand{\arraystretch}{1.5}
    \begin{tabularx}{\textwidth}{c|c|c|YYYYY|YYY}
    \toprule
         Video Backbone & Training Method & TFA & \cellcolor{yellow!25}OT (5107) & \cellcolor{yellow!35}DT (1526) & \cellcolor{orange!45}WL (297) & \cellcolor{orange!55}CT \hspace{1cm} (206) & \cellcolor{orange!65}YD \hspace{1cm} (62) & Overall mAP & Avg. C/A & {Overall Acc.} \\ \midrule 

         & CE & - & 98.2 & 91.1 & 88.0 & 56.3 & 78.0 & 82.3 & 71.8 & {90.5} \\ 
          & cRT \cite{kang2019decoupling} & - & 98.2 & 91.1 & 87.5 & 59.8 & \textbf{78.7} & 83.1 & 78.0 & {89.0} \\ 
         \textbf{I3D\tnote{*} \cite{carreira2017quo}} & Transfer-LMR (Ours) & Transformer & 98.6 & 91.3 & 86.0 & 53.9 & 77.1 & 81.4 & 79.4 & {90.2} \\
          & Transfer-LMR (Ours) & BiGRU & 98.6 & \textbf{92.2} & {86.4} & {59.2} & {72.8} & {81.9} & {80.6} & {90.4} \\ 
          \cellcolor{white}  & \textbf{Transfer-LMR (Ours)} & \textbf{Average} & \textbf{98.7} & \textbf{92.2} & \textbf{88.5} & \textbf{62.3} & 75.9 & \cellcolor{myModelColor} \textbf{83.5} & \cellcolor{myModelColor} \textbf{81.0} & \cellcolor{myModelColor} {\textbf{91.0}} \\ \hline 
         
          & CE & - & \textbf{97.9} & 86.5 & 80.1 & 32.1 & 56.9 & 70.7 & 58.8 & {88.2} \\ 
         C3D\tnote{\ddag} \cite{tran2015learning}  & cRT \cite{kang2019decoupling} & - & 97.8 & 86.5 & 80.2 & 36.3 & 48.1 & 69.8 & 64.1 & {87.3} \\ 
          \cellcolor{white}  & \textbf{Transfer-LMR (Ours)} & - & 97.4 & \textbf{86.6} & \textbf{82.2} & \textbf{38.7} & \textbf{59.3} & \textbf{72.8} & \textbf{65.1} & {\textbf{89.0}} \\ \hline

          & CE & - & 95.8 & 83.6 & 82.0 & 11.2 & \textbf{84.3} & 61.4 & 54.8 & {85.7} \\ 
         MotionFormer\tnote{\dag} \cite{patrick2021keeping} & cRT \cite{kang2019decoupling} & - & 97.3 & \textbf{89.5} & \textbf{82.6} & 23.2 & 56.8 & 69.8 & 59.5 & {89.3} \\ 
         \cellcolor{white}  & \textbf{Transfer-LMR (Ours)} & - & \textbf{97.4} & \textbf{89.5} & 80.8 & \textbf{30.3} & 61.3 & \textbf{71.9} & \textbf{63.4} & {\textbf{89.4}} \\ \hline \hline
         
         I3D\tnote{**} & CE & - & 98.5 & 90.4 & 88.4 & 48.2 & 73.9 & 79.9 & 71.6 & {90.4} \\ 
          & \textbf{Transfer-LMR (Ours)} & \textbf{Average} & \textbf{98.7} & \textbf{92.2} & \textbf{89.7} & \textbf{60.3} & \textbf{74.9} & \textbf{83.2} & \textbf{81.0} & {\textbf{91.0}} \\
        
    \bottomrule
    \end{tabularx}
    \begin{tablenotes}[para,flushleft]
        \item[\dag] Pre-Trained on EpickKitchens-100 \cite{damen2020rescaling} \quad \quad \quad \quad \quad
        \item[\ddag] Pre-Trained on UCF101 \cite{soomro2012ucf101} \quad  \quad \quad \quad \quad
        \item[*] Pre-Trained on Kinteics-400 \cite{carreira2017quo} \quad \quad \quad \quad \quad
        \item[**] Pre-Trained on HDD \cite{ramanishka2018toward}
    \end{tablenotes}
    \end{threeparttable}
    \end{center}
\end{table*}

\section{Experiments}
\subsection{Datasets}

We perform extensive experiments on two largest publicly available egocentric driving video datasets: METEOR \cite{chandra2023meteor} and HDD \cite{ramanishka2018toward}. These datasets include diverse, atypical, and unique driving behaviors captured in different road and traffic conditions. 
 
\textbf{METEOR}: The METEOR dataset \cite{chandra2023meteor} contains 1250 one-minute front-view driving videos having 1920$\times$1080 pixels resolution captured at 30fps in dense, heterogeneous, and unstructured traffic situations of Hyderabad City, India. It includes five unique atypical ego-driving behaviors with a heavy tail distribution. We follow \cite{chandra2023meteor} and split the METEOR dataset into 1000 videos for training and 231 videos for testing. Given the frame-level annotations in the dataset, we extracted 7198 trimmed intervals of driving behavior video clips from the training set and 1808 such videos from the test set. The number of such videos per class ranges from 62 to 5107 in the training set and the average duration of these videos is 1.8 seconds. 

\textbf{HDD}: This dataset includes 104 hours of 137 driving video sessions with frame-level annotations of 11 different goal-oriented tactical driving behaviors which are observed in sparse and structured traffic environments. The videos have a resolution of 1280$\times$720 captured at 30fps. We followed prior work \cite{ramanishka2018toward}\cite{li2020learning} and used 100 sessions for training and 37 sessions for testing and extracted 9730 trimmed driving behavior videos from the training set and 2647 such videos from the test set. The number of such videos per class ranges from 68 to 4859 in the training set and the average duration of these videos is 3.7 seconds.

\subsection{Evaluation Metrics}

We follow most previous works \cite{ramanishka2018toward, li2020learning} and use overall mean average precision (Overall mAP) as an evaluation metric for estimating the model's performance on the entire dataset. We also report average precision (AP) values for each behavior category to monitor class-wise performance variations. However, to validate the effectiveness of our approach in improving the class-wise average performance, we report the average class accuracy (Avg. C/A) as our main evaluation metric. Given the imbalanced test sets of both datasets, the overall (weighted) class accuracy (Overall Acc.) is also reported for reference. All values are reported in percentages.

\subsection{Baselines}

\textbf{CE.} A standard training method with instance-balanced sampling and cross-entropy loss function. This baseline is equivalent to the first training stage of our proposed approach which is described in Section \ref{sub: twostages}. 



\textbf{cRT.} Classifier Re-Training \cite{kang2019decoupling}. Originally proposed for long-tailed image recognition, it involves training with instance-balanced sampling followed by a classifier reset and class-balanced sampling. This method was modified for its application on driving video datasets by obtaining the CE baseline first followed by classifier re-training using class-balanced sampling without resetting its weights. This method alleviates the class imbalance problem and serves as a strong baseline for comparison with our approach.


\begin{table*}[!ht]
    \begin{center}
    \begin{threeparttable}
    \caption{Driving Behavior Recognition on \textbf{HDD dataset \cite{ramanishka2018toward}}. Abbreviations for driving behavior category: Intersection Passing (IP), Left turn (LT), Right turn (RT), Crosswalk passing (CW), Left lane change (LLC), Right lane change (RLC), Left lane branch (LLB), Merge (MG), Right lane branch (RLB), Railroad passing (RP), U-turn (UT). The frequency of each behavior class in the training set is specified in round brackets alongside its abbreviation. The column color shades from yellow to orange indicate the heavy-tailed distribution of driving behavior categories in the dataset.}
     
    \label{tab:HDDresults}
    \scriptsize
    \hyphenpenalty=10000 
    \exhyphenpenalty=10000 
    \renewcommand{\arraystretch}{1.5}
    \begin{tabularx}{\textwidth}{c|c|YYYYYYYYYYY|YYY}
    \toprule
        Video Backbone & Training Method & \cellcolor{yellow!20}IP (4859) & \cellcolor{yellow!25}LT (1368) & \cellcolor{yellow!30}RT (1363) & \cellcolor{yellow!35}CW (569) & \cellcolor{yellow!40}LLC (496) & \cellcolor{yellow!45}RLC (463) & \cellcolor{orange!50}LLB (245) & \cellcolor{orange!55}MG (131) & \cellcolor{orange!60}RLB (97) & \cellcolor{orange!65}RP (71) & \cellcolor{orange!70}UT (68) & Overall mAP & Avg. C/A & {Overall Acc.} \\ \midrule
        
        & CE\tnote{\dag} \cite{li2020learning} & 85.6 & 79.1 & 78.9 & 29.8 & 74 & 62.4 & 59 & 20.1 & 14.3 & 0.1 & 41.4 & 49.5 & - & {-} \\ 
        \textbf{I3D\tnote{*} \cite{carreira2017quo}}  & CE & \textbf{99.3} & \textbf{99.0} & \textbf{99.6} & 87.4 & 92.3 & \textbf{93.8} & \textbf{91.0} & 74.0 & 65.1 & \textbf{19.5} & 71.7 & 81.1 & 77.8 & {\textbf{95.0}} \\
         \cellcolor{white}  & \textbf{Transfer-LMR Avg. (Ours)} & \textbf{99.3} & 98.2 & \textbf{99.6} & \textbf{89.1} & \textbf{93.9} & 92.9 & 90.8 & \textbf{76.6} & \textbf{65.4} & \textbf{19.5} & \textbf{86.4} & \cellcolor{myModelColor} \textbf{82.8} & \cellcolor{myModelColor} \textbf{80.7} & \cellcolor{myModelColor} {94.4} \\ \hline
        
         & CE\tnote{\dag} \cite{li2020learning} & 82.4 & 77.4 & 80.7 & 17.4 & 67.9 & 56.9 & 59.7 & 20.1 & 5.2 & 3.9 & 29.5 & 45.5 & - & {-} \\ 
        C3D\tnote{\ddag} \cite{tran2015learning}  & CE & \textbf{99.4} & \textbf{98.2} & 99.4 & 84.5 & \textbf{97.9} & \textbf{89.2} & 93.5 & \textbf{81.4} & 43.7 & 28.8 & \textbf{75.1} & \textbf{81.0} & 74.5 & {92.9} \\ 
         & \textbf{Transfer-LMR (Ours)} & 99.0 & 98.0 & \textbf{99.5} & \textbf{88.3} & 97.6 & 88.7 & \textbf{93.7} & 80.4 & \textbf{47.5} & \textbf{29.4} & 68.8 & \textbf{81.0} & \textbf{76.7} & {\textbf{93.1}} \\ \hline  \hline

        I3D\tnote{**}  & CE & \textbf{99.5} & \textbf{99.1} & \textbf{99.5} & 87.3 & 94.0 & 93.7 & 86.1 & \textbf{68.9} & 93.7 & \textbf{28.6} & 63.1 & 80.4 & 76.2 & {\textbf{94.7}} \\
         & \textbf{Transfer-LMR Avg. (Ours)} & \textbf{99.5} & 99.0 & 99.4 & \textbf{88.3} & \textbf{94.8} & \textbf{94.8} & \textbf{86.2} & 65.3 & \textbf{94.8} & 16.2 & \textbf{72.9} & \textbf{80.5} & \textbf{79.7} & {94.0} \\       
        
    \bottomrule
    \end{tabularx}
    \begin{tablenotes}[para,flushleft]
        \item[\ddag] Pre-Trained on UCF101 \cite{soomro2012ucf101} \quad \quad
        \item[*] Pre-Trained on Kinteics-400 \cite{carreira2017quo} \quad \quad
        \item[**] Pre-Trained on METEOR \cite{chandra2023meteor} \quad \quad
        \item[\dag] Trained with Standard Random-Cropping Augmentation
    \end{tablenotes}
    \end{threeparttable}
    \end{center}
\end{table*}

\subsection{Implementation Details}

\textbf{Backbones.} Our proposed training method is compared with the baselines for different video backbones such as C3D \cite{tran2015learning}, I3D \cite{carreira2017quo}, and MotionFormer \cite{patrick2021keeping} which have shown substantial success in general human action recognition tasks. We establish new baselines and benchmark our method using all the aforementioned video backbones on METEOR dataset. Following prior works \cite{li2020learning}, we benchmark C3D and I3D-based models for comparison on HDD dataset. The C3D model was pre-trained on the Sports-1M and UCF101 datasets \cite{karpathy2014large}\cite{soomro2012ucf101}, while the I3D model with ResNet-50 backbone \cite{he2016deep} was pre-trained on the Kinectics-400 human action videos dataset \cite{carreira2017quo}. The MotionFormer model with ViT backbone \cite{dosovitskiy2020image} was pre-trained on the EpicKitchens-100 dataset \cite{damen2020rescaling}. These pre-trained models were then fine-tuned on respective driving datasets. Through these experiments we showcase the generalizability of our approach across video backbones and datasets for heavy-tailed driving behavior recognition.

\textbf{Model Input.} The C3D backbone intakes 16-frame clip with 112$\times$112 pixels resolution, MotionFormer takes 16-frame clip with 224$\times$224 resolution, and I3D takes 64-frame clip with 224$\times$224 resolution as default model inputs. The clip frames are randomly sampled from the trimmed behavior videos at intervals of 3fps for HDD dataset based on the prior work \cite{li2020learning} and 30fps for METEOR dataset. A higher frame sampling rate was required to obtain a more granular context in the case of METEOR. This helped in capturing the short-lived duration of the unexpected and atypical driving behaviors. The input to LMR-based feature refinement module is the feature vector from the last layer of the video backbone with dimension 4096$\times$1 in C3D and 768$\times$1 in MotionFormer, where the temporal dimension is 1. The I3D model gives output features of dimension 2048$\times$7 from the 3D average polled \textit{Mixed\_5c} layer. These features are temporally aggregated to yield 2048$\times$1 dimensional vector before feeding it as input to the LMR module.

\textbf{Training Routine.} We perform all experiments on a system with 4$\times$3080Ti GPUs using PyTorch. In the first training stage, the MotionFormer-based model is trained for 30 epochs using Adam optimizer with a learning rate of 0.0001. The C3D and I3D-based models are trained for 40 epochs using SGD optimizer with learning rates of 0.000125 and 0.1 respectively. In the second stage for cRT and Transfer-LMR methods, the MotionFormer-based model is fine-tuned with a reduced learning rate of 0.00001 while the C3D and I3D-based models are finetuned with learning rates of 0.0000125 and 0.001 respectively. The hyperparameters for the LMR module were set to 0.4 and 1.0 for reconstruction weight and decay values respectively. For a fair comparison, we keep these hyperparameters the same for all the experiments conducted on both METEOR and HDD datasets. They can otherwise be fine-tuned for each dataset separately to obtain better results. 



\section{Results}

Table \ref{tab:METEORresults} presents the results of our newly established baselines and experiments using Transfer-LMR for heavy-tailed driving behavior recognition on the validation set of METEOR dataset. Our approach consistently outperforms baseline training methods (CE and cRT) on all dataset-wise metrics: average class accuracy, overall mAP, and overall accuracy, across all video backbones (MotionFormer, C3D, and I3D). As compared to the CE baseline, our method shows significant improvement in average class accuracy by 8.6\%, 6.3\%, and 9.2\% using MotionFormer, C3D, and I3D backbones respectively. These gains indicate that our approach mitigates the class imbalance problem by improving the per-class accuracy without sacrificing the model's overall accuracy and mAP. Specifically, our approach benefits the highly underperforming Cutting (CT) tailed class by significantly boosting its performance across baselines.



\begin{figure*}[t]
    \centering
    \includegraphics[width=\textwidth]{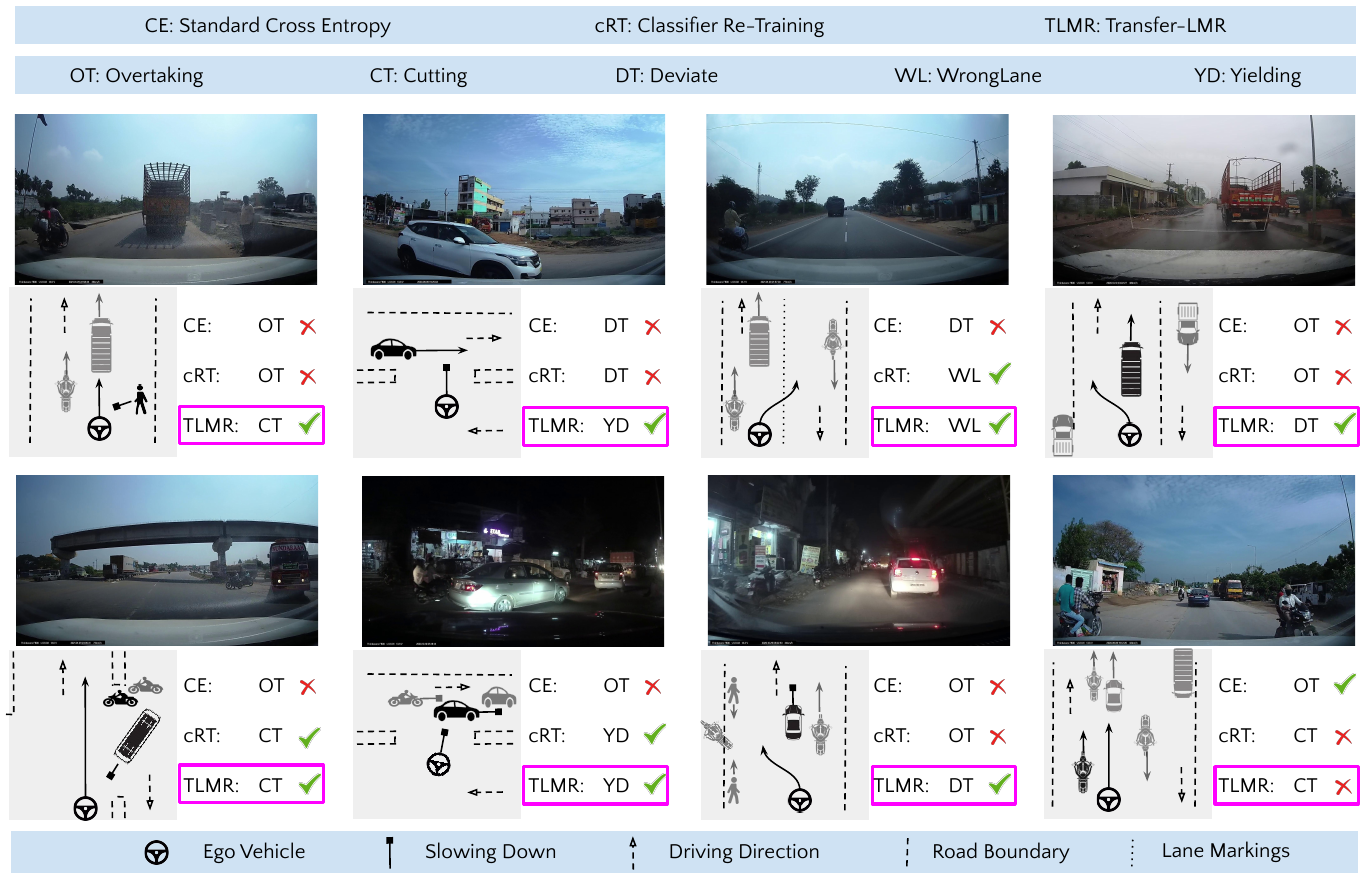}
    \caption{Qualitative Results for Driving Behavior Recognition on \textbf{METEOR dataset \cite{chandra2023meteor}}. The key frame of sample driving videos from the validation set of METEOR dataset is shown. Their Bird's Eye View (BEV) is also displayed demonstrating the vehicle trajectories that define the ego's driving behavior when observed in the video's full duration. The behavior predictions of CE and cRT baselines are compared with Transfer-LMR (TLMR) method when using the (Kinetics-400 pre-trained) I3D video backbone.}
    \label{fig: qualitative_fig_meteor}
\end{figure*}

\begin{figure*}[t]
    \centering
    \includegraphics[width=\textwidth]{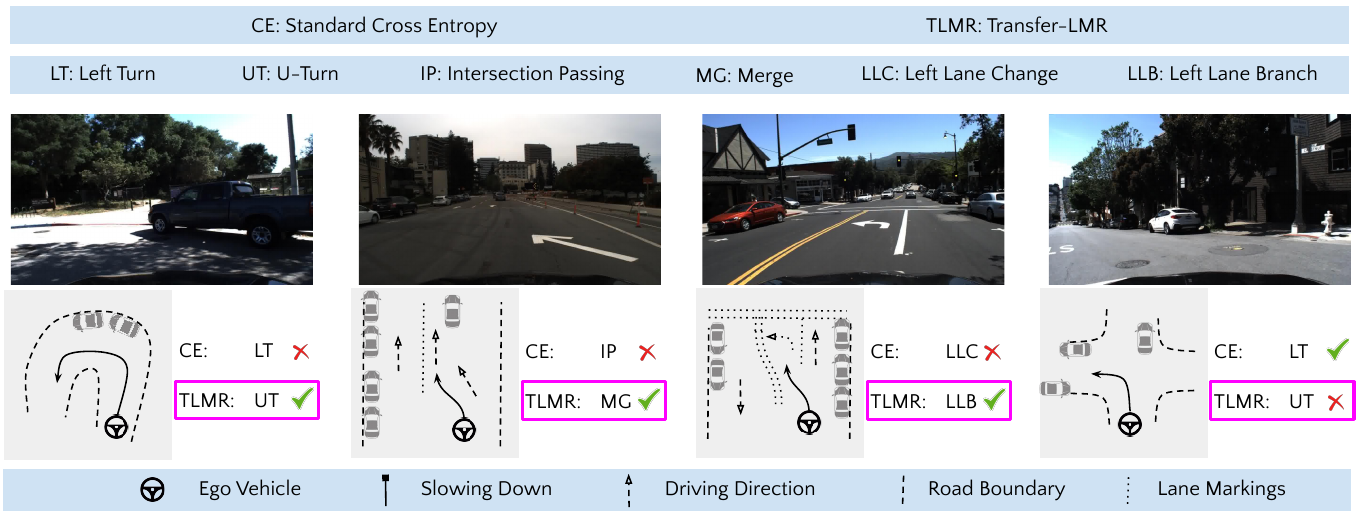}
    \caption{Qualitative Results for Driving Behavior Recognition on \textbf{HDD dataset \cite{ramanishka2018toward}}. The key frame of sample driving videos from the validation set of HDD dataset is shown. Their Bird's Eye View (BEV) is also displayed demonstrating the vehicle trajectories that define the ego's driving behavior when observed in the video's full duration. The behavior predictions of the CE baseline are compared with the Transfer-LMR (TLMR) method when using the (Kinetics-400 pre-trained) I3D video backbone.}
    \label{fig: qualitative_fig_hdd}
\end{figure*}

 
Figure \ref{fig: qualitative_fig_meteor} shows the selected examples of driving behaviors from the METEOR dataset. Qualitatively, the CE method fails to correctly identify the tail classes (CT, YD, WL, and DT). And, it gets confused among similar driving behaviors, such as, it predicts CT as OT and WL as DT class. The similarity between these behaviors is attributed to the similar trajectories followed by the ego-vehicle which can be observed in their BEVs. CT and OT differ only regarding the ego-relative orientation of the highlighted road agents in relation to which the behaviors are defined. While WL and DT behaviors differ only regarding the illegal and legal lane changes undertaken by them respectively. Also, CT being a tail class has limited diversity as compared to the head class OT, thereby it gets falsely predicted as OT with the CE method. Similarly, WL gets falsely predicted as DT due to insufficient diversity being a tail class. Transfer-LMR specifically addresses this issue by increasing the diversity among such correlated examples. This results in reduced confusion among such classes (CT, WL, DT, and OT) and substantial improvements in their average precision values over the CE baseline, as evident from Table \ref{tab:METEORresults} for the I3D-based backbone. Figure \ref{fig: qualitative_fig_meteor} also shows that our approach (TLMR) consistently gives correct predictions for the ambiguous tail classes. While a failure case shows that it incorrectly predicts the head class OT.

In Table \ref{tab:HDDresults} we benchmark and compare our proposed method with the existing method \cite{li2020learning} for driving behavior recognition on the HDD dataset. Since the modified random cropping data augmentation method, described in Section \ref{sub: D_S_A}, is employed throughout the Transfer-LMR training stages, we reproduce the CE baselines of C3D and I3D backbones for a fair comparison. We also include the existing benchmarks, reported in \cite{li2020learning}, to showcase the gap in performance as compared to our established baselines. Transfer-LMR outperforms in average class accuracy by a good margin on both the video backbones (C3D and I3D), while being comparable or better concerning overall mAP and overall accuracy. Figure \ref{fig: qualitative_fig_meteor} shows selected examples from the HDD dataset, where the CE method confuses tail classes with similar head classes (eg. LT and UT, LLC and LLB, etc. having similar ego-vehicle trajectories). Consistently, Transfer-LMR predicts the tail classes correctly while the head LT class is incorrectly predicted. Please view the video in Supplementary for additional details.


Based on the quantitative results reported in Tables \ref{tab:METEORresults} and \ref{tab:HDDresults}, we observe that I3D is a better alternative than C3D and MotionFormer for attaining superior driving behavior recognition performance. Specifically using the Kinetics-400 pre-trained I3D, Transfer-LMR gives the best results for both driving datasets METEOR and HDD. This thereby demonstrates that our training method (Transfer-LMR) helps in disambiguation features between similar driving behaviors and addresses the heavy tail problem across different driving datasets by boosting the average class accuracy and overall mAP on the entire dataset.
 



As an ablation, we perform experiments of pre-training the I3D backbone with a driving dataset, instead of any large-scale human-action videos dataset, before applying the Transfer-LMR method. This was done to observe any effects on performance if pre-training and fine-tuning dataset domains are kept similar. The results in the last two rows of Tables \ref{tab:METEORresults} and \ref{tab:HDDresults} indicate that Transfer-LMR outperforms the CE baseline significantly in average class accuracy with METEOR or HDD driving datasets being used for pre-training or fine-tuning. However, these performances were comparable but slightly inferior to the best results obtained with Kinetics-400 as the pre-training dataset. Thus it was concluded that pre-training on large-scale video datasets always helps in obtaining better performance as compared to using a similar domain and resource-constrained driving dataset for pre-training.

\section{CONCLUSIONS}

We introduce Transfer-LMR, a novel training method for heavy-tail distributed driving behaviors recognition from front-view dashcam videos. Our method alleviates the class imbalance problem in resource-constrained driving datasets by improving the class-wise average performance while maintaining better overall performance. To accomplish this we incorporate different data sampling strategies, boundary-constrained random cropping, temporal feature aggregation, and LMR-based feature refinement adapted toward modeling driving behaviors. Specifically, LMR increases the diversity of the tail classes by amalgamating similar features followed by random mixing among the training data samples. We demonstrate the effectiveness of our approach and establish state-of-the-art results for heavy-tail driving behavior recognition on METEOR and HDD datasets captured in structured and unstructured traffic environments. 









\bibliographystyle{IEEEtran}
\bibliography{IEEEfull}

\end{document}

%% file: macros.tex
\usepackage{threeparttable}

\makeatletter
\newcommand\footnoteref[1]{\protected@xdef\@thefnmark{\ref{#1}}\@footnotemark}
\makeatother

\newcommand{\shorteq}{%
  \settowidth{\@tempdima}{-}
  \resizebox{\@tempdima}{\height}{=}%
}

\usepackage[colorlinks=true, citecolor=magenta]{hyperref}
\usepackage[table]{xcolor}
\usepackage{tablefootnote} 

\definecolor{myModelColor}{rgb}{0.9,0.7,1.0}

\usepackage{amssymb,fge}

\usepackage{amsmath, amssymb}
\usepackage{pifont}
\usepackage{subcaption}
\usepackage[utf8]{inputenc}
\usepackage[english]{babel}
\usepackage[linesnumbered,ruled,vlined]{algorithm2e}
\usepackage[font=small,belowskip=-1.5pt]{caption} 
\usepackage{anyfontsize}

\usepackage{array}
\usepackage{graphicx}
\usepackage{amsfonts}
\usepackage[11pt]{moresize}
\usepackage{bm}
\usepackage{soul}
\usepackage{hhline}
\usepackage{multirow, makecell, multicol}
\usepackage{float}
\usepackage{booktabs,wrapfig}

\usepackage{amsthm}
\usepackage{color}
\usepackage{transparent}
\usepackage{url}
\usepackage{footmisc}
\usepackage{setspace}
\usepackage{mathtools}
\usepackage{gensymb}
\usepackage[export]{adjustbox}

\theoremstyle{plain}

\usepackage{graphicx}
\usepackage[T1]{fontenc}
\usepackage{pifont}
\usepackage{comment}
\usepackage{tabularx} 
\usepackage{array} 
\newcolumntype{Y}{>{\centering\arraybackslash}X}
\newcolumntype{Z}{>{\centering\arraybackslash}m{0.93cm}}

\newcolumntype{M}{>{\centering\arraybackslash}m{1.52cm}}